\documentclass[12pt,letterpaper]{article}
\usepackage[utf8]{inputenc}
\usepackage{nameref}
\usepackage[hyperfootnotes=false]{hyperref}
\usepackage{multirow}
\usepackage{tikz}
\usepackage{pgfplots,pgfplotstable}
\pgfplotsset{compat=1.14}
\usepackage{ifthen}
\usepackage{marvosym}
\usepackage{float}
\usepackage[aboveskip=1pt,labelfont=bf,labelsep=period,justification=justified,singlelinecheck=off]{caption}

\title{Shielding Google's language toxicity model against adversarial attacks}

\author{
Nestor Rodriguez\textsuperscript{$\star$}
\and 
Sergio Rojas--Galeano\textsuperscript{$\star\star$}
}
\date{}

\begin{document}

\maketitle
\phantomsection
\renewcommand{\thefootnote}{$\star$} 
\footnotetext{Independent researcher, Munich, Germany}
\renewcommand{\thefootnote}{$\star\star$} 
\footnotetext{Universidad Distrital FJC, School of Engineering, Bogotá, Colombia}
\renewcommand{\thefootnote}{} 
\footnotetext{Correspondance email: srojas\MVAt udistrital.edu.co}
\renewcommand{\thefootnote}{\arabic{footnote}} 

\vspace{-1cm}
\begin{abstract}
\hspace{-.6cm}\textbf{Background.} 
Lack of moderation in online communities enables participants to incur in personal aggression, harassment or cyberbullying, issues that have been accentuated by extremist radicalisation in the contemporary post-truth politics scenario. This kind of hostility is usually expressed by means of toxic language, profanity or abusive statements. Recently Google has developed a machine-learning-based toxicity model in an attempt to assess the hostility of a comment; unfortunately, it has been suggested that said model can be deceived by adversarial attacks that manipulate the text sequence of the comment.\\
\textbf{Methods/Results.} In this paper we firstly characterise such adversarial attacks as using obfuscation and polarity transformations. The former deceives by corrupting toxic trigger content with typographic edits, whereas the latter deceives by grammatical negation of the toxic content. Then, we propose a two--stage approach to counter--attack these anomalies, bulding upon a recently proposed text deobfuscation method and the toxicity scoring model. Lastly, we conducted an experiment with approximately 24000 distorted comments, showing how in this way it is feasible to restore toxicity of the adversarial variants, while incurring roughly on a twofold increase in processing time.\\
\textbf{Conclusions.} Even though novel adversary challenges would keep coming up derived from the versatile nature of written language, we anticipate that techniques combining machine learning and text pattern recognition methods, each one targeting different layers of linguistic features, would be needed to achieve robust detection of toxic language, thus fostering aggression--free digital interaction.
\end{abstract}

\section{Introduction}
Grecian \emph{agora} was the public place where citizens in ancient times gathered to debate current affairs and exercise rhetoric as a way to persuade audiences to follow a proposal for action. Nowadays digital media such as social networks platforms, originally conceived as simple virtual cork boards to exchange information among friends, have evolved to become contemporary \emph{agoras}, where any person with an Internet--connected device may express their opinions and debate them openly and freely. 

Unfortunately, not only the medium but also the discourse have changed, and rhetoric arguments are now frequently based on emotion rather than reason, yielding discussions intended to ridicule, distort or confuse other's opinion, avoiding factual--based debate in favour of opinion manipulation by means of fake news, libel and personal or social group hostility, an scenario now commonly--referred as post--truth politics \cite{Dale2017}.

Emotion--guided arguments may lead easily to radicalism in political, religious, ethnic, sport or minorities views, which in turn may result in comments coloured with personal aggression, harassment or cyberbullying \cite{Hosseinmardi2016, Burnap2016, Nobata2016}. This kind of hostility is becoming a cause of concern in online communities; hence, automatic moderation tools are needed to prevent caustic behaviour within the steadily increasing massive amounts of daily discussions generated in social media. In this direction, Google Counter-Abuse Technology Team has launched  \emph{Perspective}, a tool to identify toxicity of a written comment based on crowd--sourcing and machine learning models trained on large datasets of toxic conversations, as an attempt to provide safer places for online discussions \cite{Perspective2017}. 

Despite the remarkable efficacy of this tool to identify high--calibre language in diverse hot topics such as US Presidential election, Brexit and climate change, it has been suggested recently that its detection mechanism can be heavily defeated using adversarial strategies that corrupt the input text sequence with typographic or polarity manipulation, to such a degree that becomes unrecognisable to the trained model but remains readable by the human eye. For example, Hosseini et al. \cite{Hosseini2017} has shown that the insulting statement ``They are liberal idiots who are uneducated'' (toxicity: 90\%), becomes a mild comment when written as ``They are liberal i.diots who are un.educated'' (toxicity: 15\%). Similarly, the rude sentence ``It’s stupid and wrong'' (toxicity: 89\%), remains rude even if negated: ``It’s not stupid and wrong'' (toxicity: 83\%). Since the space of possible sequence variations obtainable with these kind of attacks is combinatorial, it would not be feasible to train a model even with large amounts of available examples. 

In this paper a counter--attack strategy is devised; it consists of firstly pre--processing the input with a recently proposed text deobfuscation method \cite{Rojas2017} so as to transform it back into its original representation, and subsequently feeding this corrected text to the toxic scoring model. Our results indicate that this approach is able to effectively restore the intended toxic score of the corrupted text given by \emph{Perspective},  contributing so to shield Google toxicity model against these sort of adversarial attacks. 


\section{Materials and Methods}
\subsection{Google toxicity model}
The Google \emph{Perspective} (GP) model aims to score toxicity of harassment comments in online platforms on a scale from 0 (``healthy'') to 1 (``very toxic'') . The model was built based on large--scale datasets of abusive comments, using crowd--sourced annotations  to train machine learning classifiers (logistic regression and neural networks,  with bag--of--words features, see \cite{Perspective2017} for details). Toxic content is defined as ``rude, disrespectful, or unreasonable comment that is likely to make you leave a discussion'' \cite{GPwebsite}.  

\subsection{Adversarial attacks}
Within the machine learning community, an adversarial framework refers to inputs deliberately designed to manipulate the expected behaviour of a prediction model \cite{Laskov2010, Samanta2017}. The adversary usually picks data from distributions different to those assumed when training the model, thus defeating its prediction capabilities. The attacks usually consist of corrupted features or distorted inputs. In our problem of interest, two adversarial attacks on the GP toxic model have been recently suggested \cite{Hosseini2017}, which we formally characterise as follows.

\textbf{Obfuscation attack.}
In this attack the adversary modifies the character sequence of those words conveying most of the toxic content within the input comment. The modification takes advantage of the robustness of the human visual system to recognise corrupted variants of text. Such edits can be: misspellings or symbol substitution, letter repetition, fake punctuation (inserting dots, commas or blanks within letters in the words). As a result, the input text becomes disguised from its original character codification. These kind of obfuscations has been identified as homoglyph substitution and bogus segmentation anomalies \cite{Rojas2017}. Using these stratagems, attacks have been reported to effectively reduce the toxicity score of an aggressive comment down to a benign level \cite{Hosseini2017}. 

\textbf{Polarity attack.}
In this attack the adversary attempts to obtain high toxicities for inoffensive comments that however exhibit profanity content. This is achieved by simply negating toxic terms that effectively swaps the polarity of the comment to its opposite meaning or emotional intent. For example, if the toxic word is an adjective, the polarity change is achieved by inserting  the particle ``\emph{not}''  before the word. The modified comment still contains the character sequence of the toxic term, misleading the model to identify it as aggressive even though its deceiving neutrality may originate from idiomatically uncommon or odd--looking sentences. This kind of vulnerability was also exploited effectively in \cite{Hosseini2017} to outwit the toxic model. We remark that the adversarial intensity can be worsen by combining the two attacks within the same input, i.e. negating obfuscated versions of toxicity terms.
 
\subsection{Dataset preparation}
The GP website \cite{GPwebsite} provides a sample of comments gathered from online surveys on three delicate topics:  US Election (45 comments), Brexit (61) and Climate change (49). The comments' text along with scores obtained with the GP toxicity model are given. So we collected those with toxicity scores higher than 60\%,  obtaining a subset of 24 comments that were labeled as  \{\texttt{usel01},\ldots, \texttt{usel10}, \texttt{brex01},\ldots, \texttt{brex07}, \texttt{clic01},\ldots, \texttt{clic07}\}. The complete list of comments is provided as supplementary information (\nameref{sup:comments}).

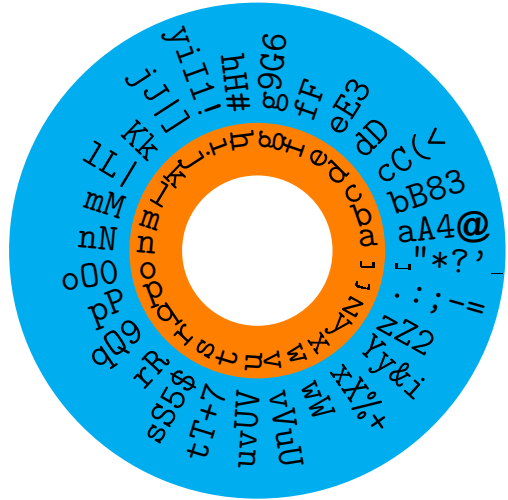
\begin{figure}[tb]
\centering
\def\delta{13} 
\def\inner{1cm} 
\def\outer{1.2cm} 

\begin{tikzpicture}[text=black,font=\ttfamily\large]
  \fill[cyan] circle(\outer+2.1cm);
  \fill[orange] circle(\outer+.5cm);
  \fill[white] circle(\inner);  
  \foreach \glyph/\list [count=\cp] in {
    a/aA4\MVAt,
    b/bB83,
    c/cC(<,
    d/dD,
    e/eE3, 
    f/fF,
    g/g9G6,
    h/hH\#,
    i/yiI1!,
    j/jJ|],
    k/Kk,
    l/lL|/\textbackslash,
    m/mM,
    n/nN,
    o/oO0,
    p/pP,
    q/qQ9,
    r/rR,
    s/sS5\$,
    t/tT+7,
    u/uvUV,
    v/vVuU,
    w/wW,
    x/xX\%+,
    y/Yy\&i,
    z/zZ2,
    \textvisiblespace/.:;-=,
    \textvisiblespace/\textvisiblespace"*?'\_,
  }{
    \ifthenelse{\equal{\glyph}{}}{}{
      \pgfmathsetmacro{\angle}{(\cp-1)*\delta + \delta/2}
      \pgfmathtruncatemacro{\anglenode}{\angle}
      \ifthenelse{\( \anglenode > 90 \) \AND \( \anglenode < 270 \)}{ 
        \node[rotate=180+\anglenode,anchor=east] at (\angle:\outer) {{\list~\glyph}};
      }{
        \node[rotate=\anglenode,anchor=west] at (\angle:\outer) {{\glyph~\list}};
      }
    }
  }
\end{tikzpicture}\\[.3cm]

\caption{Homoglyph substitution lists. Each list is indexed by the characters shown in the inner orange ring, corresponding to the letters of the English alphabet. The sets of their corresponding homoglyphs used in this study are shown in the outer blue ring. The blank character ``\textvisiblespace'' indexes the lists of bogus segmentators.}
\label{fig:homoglyphs}
\end{figure}

We subsequently prepared datasets conveying the two types of adversarial anomalies, namely obfuscation attack and polarity attack. In the first attack, 1000 variations of each comment were randomly generated by isolating the toxic terms of the comment in order to obfuscate them with a number of edits including homoglyph substitution, bogus segmentation and letter repetition (using the same scheme as \cite{Rojas2017}). The number of edits applied to each term was controlled with a corruption rate $p \in \{50\%, 99\%\}$. Thus, for each character in the term a uniformly-distributed random number $r \sim {\cal U}(0,1)$ was sampled; if $r < p$ the character was kept unchanged, otherwise one of these edits was applied (with their respective probabilities): an homoglyph substitution (60\%), a bogus segmentation (30\%) or a fake letter repetition (10\%). In this way, we obtained two datasets, \texttt{obfuscation-50} and \texttt{obfuscation-99} consisting of 24,000 obfuscated toxic comments each (see examples in Table \ref{tab:dataset}). With respect to the homoglyph edits utilised in the obfuscation attack, we defined substitution lists encompassing a subset of the ASCII encoding set, as  depicted in Fig. \ref{fig:homoglyphs}. We note in passing that these lists can be customised to further include additional homoglyphs from extended character encodings (e.g. letters with diacritical marks).

\renewcommand{\arraystretch}{1.1}
\begin{table}[tb]
\footnotesize
\caption{{\bf An excerpt of adversarial examples from the datasets.}}
\begin{tabular}{|c||c|l|p{8.5cm}|}
\hline
Topic & Attack & ID & Text sequence \\ 
\hline\hline
 & None (original) & \texttt{usel01} &	If they voted for Hilary they are idiots \\\cline{2-4}
US & \texttt{obfuscation-50} & \texttt{usel01-098} &	If they voted for Hilary they are !d10ts \\\cline{2-4}
Election & \texttt{obfuscation-99} & \texttt{usel01-295} &	If they voted for Hilary they are i;;d-I0t\_5 \\\cline{2-4}
& \texttt{polarity} & \texttt{usel01'} &	If they voted for Hilary they are NOT idiots \\\hline\hline
\multirow{4}{*}{Brexit} & None (original) & \texttt{brex03} &	They are stupid and ignorant with no class \\\cline{2-4}
& \texttt{obfuscation-50} & \texttt{brex03-381} &	They are st.VPId and ignor\MVAt Nt with no class \\\cline{2-4}
& \texttt{obfuscation-99} & \texttt{brex03-120} &	They are s7uuupi-d,, and 1g,Nooooora,n~t with no class 
\\\cline{2-4}
& \texttt{polarity} & \texttt{brex03'} &	They are NOT stupid and NOT ignorant with no class \\\hline\hline
 & None (original) & \texttt{clic01} &	They have their heads up their ass \\\cline{2-4}
Climate & \texttt{obfuscation-50} & \texttt{clic01-688} & They have their heads up their AS5 \\\cline{2-4}
change & \texttt{obfuscation-99} & \texttt{clic01-512} & They have their heads up their a**\$s \\\cline{2-4}
& \texttt{polarity} & \texttt{clic01'} &	They have their heads NOT up their ass. \\\hline

\end{tabular}
\label{tab:dataset}
\end{table} 
 
Now, the second dataset (\texttt{polarity}) was obtained by inserting negation predicates within each comment. Thus, the size of this dataset is 24 negated comments. Although possible, we refrained from generating obfuscated variants of these comments as  the aim of this attack was to investigate the effect of the polarity change achieved by the negated predicates alone. Table \ref{tab:dataset} also shows examples for this attack. The complete datasets are provided as  supplementary files (see \nameref{sup:datasets}). 

On the other hand, the list of toxic terms was build as follows. For each comment in the original dataset, we scanned every single word with the toxicity model; those scoring higher than 50\% were added to the list. Regarding the negated predicates, such list was built by preceding each word within the toxic term list, with the particle \emph{not} if it was an adjective or noun, or \emph{do not} if it was a verb. The resulting lists are shown in Table \ref{tab:lists}.

\begin{table}[!t]
 \caption{\bf The lists of toxic terms and their negated predicates. }
 \begin{tabular}{|c||c||c||c|} 
 \hline
 Toxic term & Negated predicate & Toxic term & Negated predicate \\ [0.5ex] 
 \hline\hline
idiots & not idiots & awful & not awful \\
stupid & not stupid & in hell & not in hell \\
the worst & not the worst & morons & not morons \\
screw you & don't screw you & terrible & not terrible \\
suck & don't suck & a shame & not a shame \\
stupidity & no stupidity & racist & not racist \\
sexist & not sexist & supremacists & not supremacists \\
assholes & not assholes & ignorant & not ignorant \\
a moron & not a moron & nationalists & not nationalists \\
rubbish & not rubbish & up their ass & not up their ass \\
uneducated & not uneducated & an idiot & not an idiot \\
an abortion & not an abortion && \\[.5ex] 
 \hline
 \end{tabular}
 \label{tab:lists}
\end{table}

\subsection{Disarming the adversarial attacks}
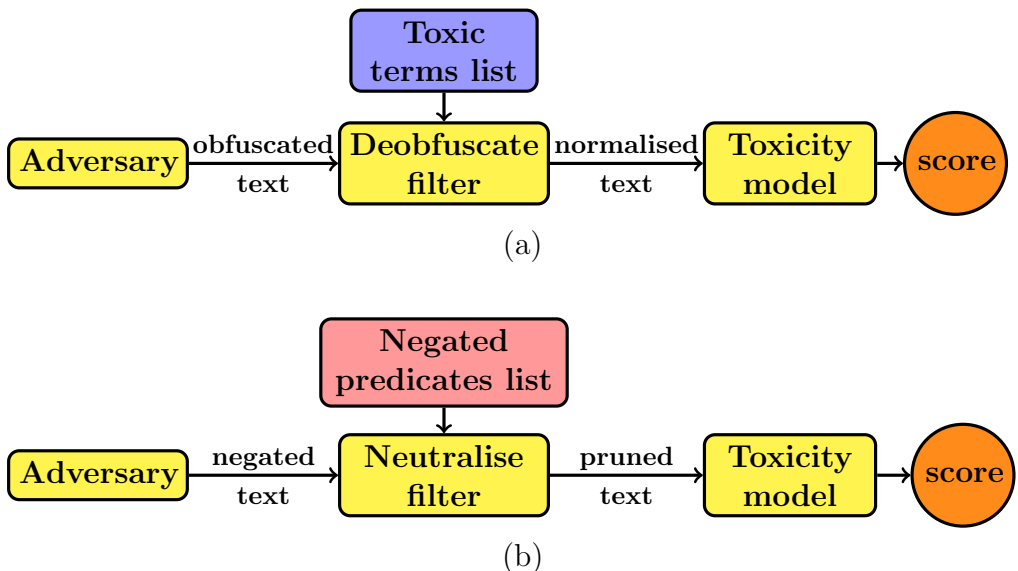
\begin{figure}[!b]
\tikzstyle{format} = [draw, fill=yellow!80]
\begin{tikzpicture}[node distance=4.6cm, align=center, text width=2.2cm, font=\bf, very thick, rounded corners]
    \path[->] node[format, text width=2.1cm] (adv) {Adversary};
    \path[->] node[format, text width=2.5cm, right of=adv] (deobf) {Deobfuscate filter}
                  (adv) edge node {\footnotesize obfuscated text} (deobf);
    \path[->] node[format, right of=deobf, text width=2cm] (tox) {Toxicity model}
                  node[node distance=1.5cm, draw, fill=blue!40, above of=deobf] (list) {Toxic terms list} edge (deobf) (deobf) edge node {\footnotesize normalised text} (tox);
    \path[->] node[node distance=2.2cm, text width=.8cm, circle, format, right of=tox, fill=orange!90, text width=1cm] (s) {score} (tox) edge (s);
\end{tikzpicture}
\centerline{{(a)}}\\[.7cm]
\begin{tikzpicture}[node distance=4.6cm, align=center, text width=2.2cm, font=\bf, very thick, rounded corners]
    \path[->] node[format, text width=2.1cm] (adv) {Adversary};
    \path[->] node[format, text width=2.5cm, right of=adv] (deobf) {Neutralise filter}
                  (adv) edge node {\footnotesize negated text} (deobf);
    \path[->] node[format, right of=deobf, text width=2cm] (tox) {Toxicity model}
                  node[node distance=1.5cm, draw, fill=red!40, text width=3cm, above of=deobf] (list) {Negated predicates list} edge (deobf) 
                  (deobf) edge node {\footnotesize pruned\\ text} (tox);
    \path[->] node[node distance=2.3cm, text width=.8cm, circle, format, right of=tox, fill=orange!90, text width=1cm] (s) {score} (tox) edge (s);
\end{tikzpicture}
\centerline{{(b)}}
\caption{{Disarming the adversarial attacks.} The adversary comment is preprocessed in order to provide a suitable input to the toxicity model. (a) Obfuscation attack: the preprocessor is a deobfuscation filter fed with a list of toxic terms; if any obfuscated variant is found, the filter replaces them with their plain English version. (b) Polarity attack: the preprocessor is a neutralisation filter fed with a list of negation predicates; if any  of these are found, in this case the filter removes them from the input text.}
\label{fig:method}
\end{figure}

The method we propose to address the adversarial attacks is illustrated in Fig. \ref{fig:method}; it basically consist of filtering the corrupted comment prior to feeding the toxicity model, by means of the deobfuscation approach introduced in \cite{Rojas2017}. Such normalisation filter requires a list of target vocables written in plain English, that are to be searched within the input text. The approach then uses an approximate string matching algorithm to find occurrences of the vocables, either verbatim copies or variants resulting from the homoglyph  substitution, bogus segmentation or symbol repetition. Matching occurrences are corrected and the resulting sequence is accordingly fed to the GP toxicity model.  

We defined two correction schemes, namely a \emph{deobfuscation} filter or a \emph{neutralisation} filter, depending on the kind of attack, obfuscation or polarity, respectively. In the former, the vocabulary consists of a list of toxic terms that when obfuscated are not recognised by the  model, thus lowering the toxicity of the comment whilst still conveying its aggressive tone; therefore here the filter corrects the input by replacing obfuscated occurrences with their corresponding plain English versions. In contrast, the neutralisation filter uses as vocabulary a list of negated predicates of such toxic terms; here  the input is corrected by removing (i.e. pruning) plain or even obfuscated occurrences of these predicates, effectively switching the polarity of the comment towards a neutral attitude by expurgating  references to (negated) toxic content. These two vocabularies are shown in Table \ref{tab:lists}.

\section{Results}
\subsection{Experimental setup}
In order to apply the toxicity model and the deobfuscation filter we used the Perspective Web API \cite{GPwebsite} and the TextPatrol Web API \cite{TPwebsite} respectively. For this purpose we developed a command-line tool that takes as input a dataset of comments that are processed by invoking the APIs in order to produce an output text file  including, for each comment, toxicity scores and execution times of the GP model alone and the combined TP+GP method. The command-line tool was written using the Go programming language, it is open source and publicly available in the following Gitlab repository \url{https://gitlab.com/textpatrol/gp-tp-experiment}, which also hosts the obfuscation and polarity attacks datasets, the set of result files and instructions on how to build the tool and re-run the experiments.

\subsection{Obfuscation attack}
Let us examine first the results of toxicity scores reported in Fig. \ref{fig:obfuscation}. The figure comprises three plots corresponding to the three topics: \texttt{usel*}, \texttt{brex*} and \texttt{clic*}. Each plot in turn combines  the scores obtained in both datasets, \texttt{obfuscation-50} and \texttt{obfuscation-99}, within two reflected panels (upper and lower half, respectively; notice that the y-axis scale ranges from 0 to 1 in both directions). Three sets of values are depicted in these plots: toxicity of the original comments (red squares) and average toxicities of the 1000 variants of each comment, as obtained by the GP model (green bars) and by the proposed TP+GP method (amber bars, whiskers being standard deviations).

\definecolor{myblue}{HTML}{4F81BD}
\definecolor{myred}{HTML}{C0504D}
\definecolor{mygreen}{HTML}{9BBB59}
\definecolor{myamber}{HTML}{FFBF00}

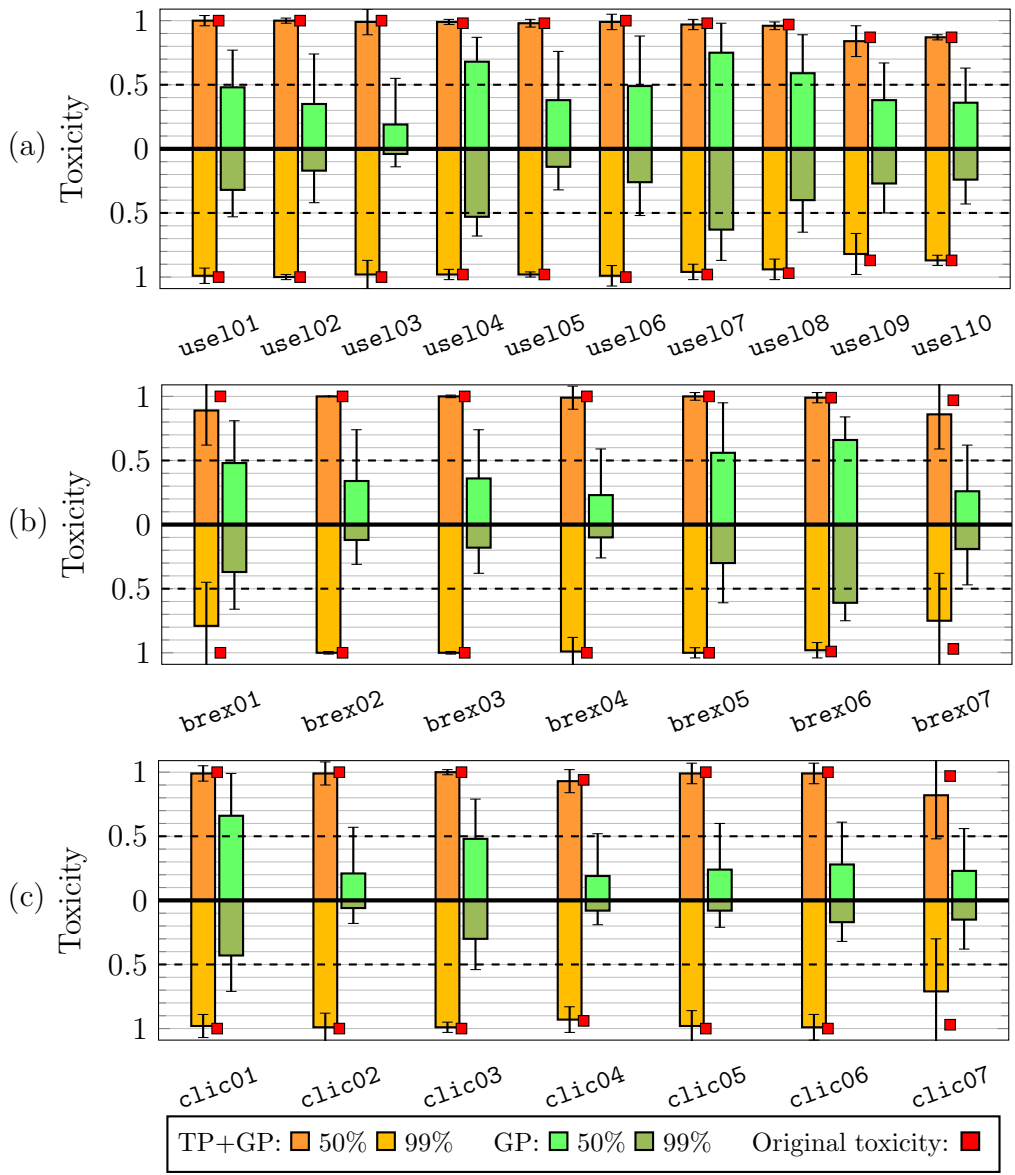
\begin{figure}[H]

\begin{tikzpicture}
    \begin{axis}[
        title = (a),
        title style = {at={(-.12,0.45)},anchor=east},
        width  = 0.94*\textwidth,
        height = 5.3cm,
        major x tick style = transparent,
        minor x tick style = transparent,
        ybar=2*\pgflinewidth,
        bar width=9pt,
        ymajorgrids = true,
        yminorgrids = true,
        minor tick num=4,
        ymin=-1.09, ymax=1.09,
        yticklabel pos=left,
        yticklabels={dummy,1,0.5,0,0.5,1},
        ylabel = {Toxicity},
        ylabel near ticks,
        symbolic x coords = {usel01,usel02,usel03,usel04,usel05,usel06,usel07,usel08,usel09,usel10},
        xticklabel style = {
            rotate=20,            font={\ttfamily\footnotesize}
        },
        enlarge x limits=0.08,
    ]
    
        \addplot[forget plot, style={black,thick,fill=orange!80,mark=none},error bars/.cd,y dir=both,y explicit, error bar style={black,thick}]
             coordinates {
             (usel01,1.00) +- (0.0, 0.04) 
             (usel02,1.00) +- (0.0, 0.02) 
             (usel03,0.99) +- (0.0, 0.10) 
             (usel04,0.99) +- (0.0, 0.02) 
             (usel05,0.98) +- (0.0, 0.03) 
             (usel06,0.99) +- (0.0, 0.06) 
             (usel07,0.97) +- (0.0, 0.04) 
             (usel08,0.96) +- (0.0, 0.03) 
             (usel09,0.84) +- (0.0, 0.12) 
             (usel10,0.87) +- (0.0, 0.02) 
            };

        \addplot[style={black,thick, fill=myamber,mark=none},error bars/.cd,y dir=both,y explicit, error bar style={black,thick}]
             coordinates {
             (usel01,-0.99) +- (0.0, 0.06) 
             (usel02,-1.00) +- (0.0, 0.02) 
             (usel03,-0.98) +- (0.0, 0.11) 
             (usel04,-0.98) +- (0.0, 0.04) 
             (usel05,-0.98) +- (0.0, 0.02) 
             (usel06,-0.99) +- (0.0, 0.08) 
             (usel07,-0.96) +- (0.0, 0.06) 
             (usel08,-0.94) +- (0.0, 0.08) 
             (usel09,-0.82) +- (0.0, 0.16) 
             (usel10,-0.87) +- (0.0, 0.04)    
            };

        \addplot[forget plot, style={black,thick,fill=green!60,mark=none},error bars/.cd,y dir=plus,y explicit, error bar style={black,thick}]
             coordinates {
             (usel01,0.48) +- (0.0, 0.29) 
             (usel02,0.35) +- (0.0, 0.39) 
             (usel03,0.19) +- (0.0, 0.36) 
             (usel04,0.68) +- (0.0, 0.19) 
             (usel05,0.38) +- (0.0, 0.38) 
             (usel06,0.49) +- (0.0, 0.39) 
             (usel07,0.75) +- (0.0, 0.23) 
             (usel08,0.59) +- (0.0, 0.30) 
             (usel09,0.38) +- (0.0, 0.29) 
             (usel10,0.36) +- (0.0, 0.27) 
             };
        
        \addplot[style={black,thick,fill=mygreen,mark=none},error bars/.cd,y dir=minus,y explicit, error bar style={black,thick}]
             coordinates {
             (usel01,-0.32) +- (0.0, 0.21) 
             (usel02,-0.17) +- (0.0, 0.25) 
             (usel03,-0.04) +- (0.0, 0.10) 
             (usel04,-0.53) +- (0.0, 0.15) 
             (usel05,-0.14) +- (0.0, 0.18) 
             (usel06,-0.26) +- (0.0, 0.26) 
             (usel07,-0.63) +- (0.0, 0.24) 
             (usel08,-0.40) +- (0.0, 0.25) 
             (usel09,-0.27) +- (0.0, 0.23) 
             (usel10,-0.24) +- (0.0, 0.19) 
             }; 
 
        \addplot[only marks,mark=square*,mark options={fill=red}] table[x=x,y=y] {
        x   y
        usel01   1.00
        usel02   1.00
        usel03   1.00
        usel04   0.98
        usel05   0.98
        usel06   1.00
        usel07   0.98
        usel08   0.97
        usel09   0.87
        usel10   0.87
        usel01   -1.00
        usel02   -1.00
        usel03   -1.00
        usel04   -0.98
        usel05   -0.98
        usel06   -1.00
        usel07   -0.98
        usel08   -0.97
        usel09   -0.87
        usel10   -0.87
        };
        
        \draw [black, ultra thick] ({rel axis cs:0,0}|-{axis cs:usel01, 0}) -- ({rel axis cs:1,0}|-{axis cs:usel10, 0});
        \draw [black, thick, dashed] ({rel axis cs:0,0.5}|-{axis cs:usel01, 0.5}) -- ({rel axis cs:1,0.5}|-{axis cs:usel10, 0.5});
        \draw [black, thick, dashed] ({rel axis cs:0,-0.5}|-{axis cs:usel01, -0.5}) -- ({rel axis cs:1,-0.5}|-{axis cs:usel10, -0.5});        
    \end{axis}
\end{tikzpicture}

\begin{tikzpicture}
    \begin{axis}[
        title = (b),
        title style = {at={(-.12,0.45)},anchor=east},
        width  = 0.94*\textwidth,
        height = 5.3cm,
        major x tick style = transparent,
        minor x tick style = transparent,
        ybar=2*\pgflinewidth,
        bar width=9pt,
        ymajorgrids = true,
        yminorgrids = true,
        minor tick num=4,
        ymin=-1.09, ymax=1.09,
        yticklabel pos=left,
        yticklabels={dummy,1,0.5,0,0.5,1},
        ylabel = {Toxicity},
        ylabel near ticks,
        symbolic x coords = {brex01,brex02,brex03,brex04,brex05,brex06,brex07},
        xticklabel style = {
            rotate=20,            font={\ttfamily\footnotesize}
        },
        enlarge x limits=0.08,
    ]
    
        \addplot[forget plot, style={black,thick,fill=orange!80,mark=none},error bars/.cd,y dir=both,y explicit, error bar style={black,thick}]
             coordinates {
             (brex01,0.89) +- (0.0, 0.27) 
             (brex02,1.00) +- (0.0, 0.00) 
             (brex03,1.00) +- (0.0, 0.01) 
             (brex04,0.99) +- (0.0, 0.09) 
             (brex05,1.00) +- (0.0, 0.03) 
             (brex06,0.99) +- (0.0, 0.04) 
             (brex07,0.86) +- (0.0, 0.27) 
            };

        \addplot[style={black,thick, fill=myamber,mark=none},error bars/.cd,y dir=both,y explicit, error bar style={black,thick}]
             coordinates {
             (brex01,-0.79) +- (0.0, 0.34) 
             (brex02,-1.00) +- (0.0, 0.01) 
             (brex03,-1.00) +- (0.0, 0.01) 
             (brex04,-0.99) +- (0.0, 0.11) 
             (brex05,-1.00) +- (0.0, 0.04) 
             (brex06,-0.98) +- (0.0, 0.06) 
             (brex07,-0.75) +- (0.0, 0.37) 
            };            

        \addplot[forget plot, style={black,thick,fill=green!60,mark=none},error bars/.cd,y dir=plus,y explicit, error bar style={black,thick}]
             coordinates {
             (brex01,0.48) +- (0.0, 0.33) 
             (brex02,0.34) +- (0.0, 0.40) 
             (brex03,0.36) +- (0.0, 0.38) 
             (brex04,0.23) +- (0.0, 0.36) 
             (brex05,0.56) +- (0.0, 0.39) 
             (brex06,0.66) +- (0.0, 0.18) 
             (brex07,0.26) +- (0.0, 0.36) 
             };

        \addplot[style={black,thick,fill=mygreen,mark=none},error bars/.cd,y dir=minus,y explicit, error bar style={black,thick}]
             coordinates {
             (brex01,-0.37) +- (0.0, 0.29) 
             (brex02,-0.12) +- (0.0, 0.19) 
             (brex03,-0.18) +- (0.0, 0.20) 
             (brex04,-0.10) +- (0.0, 0.16) 
             (brex05,-0.30) +- (0.0, 0.31) 
             (brex06,-0.61) +- (0.0, 0.14) 
             (brex07,-0.19) +- (0.0, 0.28) 
             }; 

        \addplot[only marks,mark=square*,mark options={fill=red}] table[x=x,y=y] {
        x   y
        brex01   1.00
        brex02   1.00
        brex03   1.00
        brex04   1.00
        brex05   1.00
        brex06   0.99
        brex07   0.97
        brex01   -1.00
        brex02   -1.00
        brex03   -1.00
        brex04   -1.00
        brex05   -1.00
        brex06   -0.99
        brex07   -0.97
        };
        
        \draw [black, ultra thick] ({rel axis cs:0,0}|-{axis cs:brex01, 0}) -- ({rel axis cs:1,0}|-{axis cs:brex07, 0});
        \draw [black, thick, dashed] ({rel axis cs:0,0.5}|-{axis cs:brex01, 0.5}) -- ({rel axis cs:1,0.5}|-{axis cs:brex07, 0.5});
        \draw [black, thick, dashed] ({rel axis cs:0,-0.5}|-{axis cs:brex01, -0.5}) -- ({rel axis cs:1,-0.5}|-{axis cs:brex07, -0.5});        
    \end{axis}
\end{tikzpicture}

\begin{tikzpicture}
    \begin{axis}[
        title = (c),
        title style = {at={(-.12,0.45)},anchor=east},
        width  = 0.94*\textwidth,
        height = 5.3cm,
        major x tick style = transparent,
        minor x tick style = transparent,
        ybar=2*\pgflinewidth,
        bar width=9pt,
        ymajorgrids = true,
        yminorgrids = true,
        minor tick num=4,
        ymin=-1.09, ymax=1.09,
        yticklabel pos=left,
        yticklabels={dummy,1,0.5,0,0.5,1},
        ylabel = {Toxicity},
        ylabel near ticks,
        symbolic x coords = {clic01,clic02,clic03,clic04,clic05,clic06,clic07},
        xticklabel style = {
            rotate=20,            font={\ttfamily\footnotesize}
        },
        enlarge x limits=0.08,
        name=clic
    ]
    
        \addplot[forget plot, style={black,thick,fill=orange!80,mark=none},error bars/.cd,y dir=both,y explicit, error bar style={black,thick}]
             coordinates {
             (clic01,0.99) +- (0.0, 0.06) 
             (clic02,0.99) +- (0.0, 0.09) 
             (clic03,1.00) +- (0.0, 0.02) 
             (clic04,0.93) +- (0.0, 0.09) 
             (clic05,0.99) +- (0.0, 0.08) 
             (clic06,0.99) +- (0.0, 0.08) 
             (clic07,0.82) +- (0.0, 0.34) 
            };

        \addplot[style={black,thick, fill=myamber,mark=none},error bars/.cd,y dir=both,y explicit, error bar style={black,thick}]
             coordinates {
             (clic01,-0.98) +- (0.0, 0.09) 
             (clic02,-0.99) +- (0.0, 0.11) 
             (clic03,-0.99) +- (0.0, 0.04) 
             (clic04,-0.93) +- (0.0, 0.10) 
             (clic05,-0.98) +- (0.0, 0.12) 
             (clic06,-0.99) +- (0.0, 0.10) 
             (clic07,-0.71) +- (0.0, 0.41) 
            };            

        \addplot[forget plot, style={black,thick,fill=green!60,mark=none},error bars/.cd,y dir=plus,y explicit, error bar style={black,thick}]
             coordinates {
             (clic01,0.66) +- (0.0, 0.33) 
             (clic02,0.21) +- (0.0, 0.36) 
             (clic03,0.48) +- (0.0, 0.31) 
             (clic04,0.19) +- (0.0, 0.33) 
             (clic05,0.24) +- (0.0, 0.36) 
             (clic06,0.28) +- (0.0, 0.33) 
             (clic07,0.23) +- (0.0, 0.33) 
             };

        \addplot[style={black,thick,fill=mygreen,mark=none},error bars/.cd,y dir=minus,y explicit, error bar style={black,thick}]
             coordinates {
             (clic01,-0.43) +- (0.0, 0.28) 
             (clic02,-0.06) +- (0.0, 0.12) 
             (clic03,-0.30) +- (0.0, 0.24) 
             (clic04,-0.08) +- (0.0, 0.11) 
             (clic05,-0.08) +- (0.0, 0.13) 
             (clic06,-0.17) +- (0.0, 0.15) 
             (clic07,-0.15) +- (0.0, 0.23) 
             }; 

        \addplot[only marks,mark=square*,mark options={fill=red}] table[x=x,y=y] {
        x   y
        clic01   1.00
        clic02   1.00
        clic03   1.00
        clic04   0.94
        clic05   1.00
        clic06   1.00
        clic07   0.97
        clic01   -1.00
        clic02   -1.00
        clic03   -1.00
        clic04   -0.94
        clic05   -1.00
        clic06   -1.00
        clic07   -0.97
        };

        \draw [black, ultra thick] ({rel axis cs:0,0}|-{axis cs:clic01, 0}) -- ({rel axis cs:1,0}|-{axis cs:clic07, 0});
        \draw [black, thick, dashed] ({rel axis cs:0,0.5}|-{axis cs:clic01, 0.5}) -- ({rel axis cs:1,0.5}|-{axis cs:clic07, 0.5});
        \draw [black, thick, dashed] ({rel axis cs:0,-0.5}|-{axis cs:clic01, -0.5}) -- ({rel axis cs:1,-0.5}|-{axis cs:clic07, -0.5});                
    \end{axis}

\node[draw=black,thick,rounded corners=0pt,below=10mm,font=\footnotesize] at (clic.south) {
\begin{tabular}{@{}r@{ }l@{ }l@{ }cr@{ }l@{ }l@{ }cr@{ }l@{ }}
 
 TP+GP: & 
 \tikz{\node[draw=black,thick,fill=orange!80] (0,0){};} 50\% & \tikz{\node[draw=black,thick,fill=myamber] (0,0){};} 99\% &&
 GP: &
 \tikz{\node[draw=black,thick,fill=green!60] (0,0){};} 50\% & \tikz{\node[draw=black,thick,fill=mygreen] (0,0){};} 99\% &&
 Original toxicity: &
 \tikz{\node[draw=black,thick,fill=red] (0,0){};}
\end{tabular}};

\end{tikzpicture}
\vspace{.2cm}
\caption{Toxicity scores for the obfuscation attack. In each plot the upper half panel shows average toxicities obtained for comments obfuscated with a 50\% corruption rate (\texttt{obfuscation-50} dataset) whereas the reflected lower half panel reports toxicities with a 99\% corruption rate (\texttt{obfuscation-99} dataset). Bars indicate the average was taken over 1000 variants in their respective comment category (whiskers being standard deviations) whilst squares indicate the toxicity values of a single observation, i.e. those of the original comments. (a) US Election. (b) Brexit. (c) Climate change. }
\label{fig:obfuscation}
\end{figure}

We initially note that the toxicity of the original deobfuscated comments (red squares) is nearly 1.0 for every comment, except \texttt{usel09} and \texttt{usel10} scoring toxicities closer to 0.9. These two categories contain toxic terms related to superlatives or adjective derivations (\emph{racist, supremacist, sexist, stupidity}) whereas the remainder comments incorporate either insults or obscenity (\emph{idiot, stupid, asshole, moron, screw you}) suggesting that the GP toxic model stresses profanity terms.

Next, we identify a common pattern on the results obtained by the GP model alone, namely its vulnerability to the obfuscation attack. This is no matter of surprise as this weakness in the model was previously suggested in the   preliminary study of \cite{Hosseini2017} using a few examples of the attack. Here it can be seen that in average, the GP model (green bars) decreases toxicities to levels closer to or lower than 0.5, which can be considered a moderate aggressiveness cut--off to decide if a comment is safe (18 out of 24 cases are below this cut--off in  \texttt{obfuscation-50} and 22/24 in \texttt{obfuscation-99}). A more stringent cut--off of 0.75 would have resulted in all but one of the toxic obfuscated categories been scored as non--toxic. Altogether, these findings substantiate on a large--scale basis that the adversarial obfuscation attack is effectively able to deceive the GP model.

On the other hand, we observe that in average the TP+GP method (amber bars) manages to restore the toxic scores close to their original values. On the \texttt{obfuscation-50} dataset the differences between the TP+GP scores and the original toxicity differ in less than 0.03 in all but three cases: \texttt{brex01} (0.11),  \texttt{brex07} (0.11) and  \texttt{clic07} (0.15). Similarly, on the more deceitful \texttt{obfuscation-99} dataset, where essentially the entire sequence of toxic terms occurrences are obfuscated, the reduction in the scores is again less than 0.03 in all but four cases: \texttt{usel09} (0.05), \texttt{brex01} (0.21),  \texttt{brex07} (0.22) and  \texttt{clic07} (0.26). Notice that even with a stringent 0.7 cut--off, the proposed method in average will  correctly identify all the cases as toxic. Besides, the evidence indicate that also in all cases, the average drop of toxicity incurred by the GP model is significantly larger than that the average drop obtained by the TP+GP method ($p<0.001$).

In order to further compare the effectiveness of the proposed method in disarming the obfuscation attack, we illustrate in Fig. \ref{fig:polar-hist} radial frequency histograms of the proportion of obfuscated instances correctly scored with at least the original toxicity, from the total 1000 comments (red area) in each category. It can be seen how on the \texttt{obfuscation-50} dataset the TP+GP method (amber area) is able to correctly score a large proportion of comments in all topics. This pattern is replicated in the \texttt{obfuscation-99} dataset, although proportions slightly decline due to the higher corruption level, which yields more acid attacks.

\tikzset{
  hist 1/.style={fill=mygreen},
  hist 2/.style={fill=myamber},
  hist 3/.style={fill=myred},
}

\def\delta{15} 
\def\inner{.1cm} 
\def\outer{1.7cm} 

\begin{figure}[!t]
\begin{tabular}{cc}

\multicolumn{2}{c}{
\begin{tikzpicture}
\node[draw=black,thick,below=14mm,font=\footnotesize] at (clic.south) {
\begin{tabular}{@{}r@{ }cr@{ }cr@{ }c}
 
 \tikz{\node[draw=black,thick,fill=myamber] (0,0){};} & TP+GP &
 
 \tikz{\node[draw=black,thick,fill=mygreen] (0,0){};}  & GP  &
 
 \tikz{\node[draw=black,thick,fill=myred] (0,0){};} & Total variants 
\end{tabular}};
\end{tikzpicture} 
}
\\[.3cm]

\begin{tikzpicture}[text=black,font=\ttfamily\bfseries]
  \fill[white] circle(\outer+1cm);
  \foreach \id/\counts [count=\cp] in {%
    brex03/{149,(990-149),(1000-990)},
    brex04/{157,(983-157),(1000-983)},
    brex05/{298,(991-298),(1000-991)},
    brex06/{64,(840-64),(1000-840)},
    brex07/{88,(812-88),(1000-812)},
    clic01/{377,(964-377),(1000-964)},
    clic02/{157,(990-157),(1000-990)},
    clic03/{137,(987-137),(1000-987)},
    clic04/{149,(935-149),(1000-935)},
    clic05/{169,(964-169),(1000-964)},
    clic06/{149,(935-149),(1000-935)},
    clic07/{97,(826-97),(1000-826)},
    usel01/{144,(983-144),(1000-983)},
    usel02/{148,(987-148),(1000-987)},
    usel03/{149,(940-149),(1000-940)},
    usel04/{56,(885-56),(1000-885)},
    usel05/{39,(871-39),(1000-871)},
    usel06/{318,(935-318),(1000-935)},
    usel07/{238,(845-238),(1000-845)},
    usel08/{96,(818-96),(1000-818)},
    usel09/{109,(896-109),(1000-896)},
    usel10/{46,(907-46),(1000-907)},
    brex01/{146,(827-146),(1000-827)},
    brex02/{157,(990-157),(1000-990)},
  }{
    \ifthenelse{\equal{\id}{}}{}{
      \pgfmathsetmacro{\angle}{(\cp-1)*\delta-90+\delta/2}
      \pgfmathsetmacro{\total}{\inner}
      \pgfmathsetmacro{\am}{\outer-\inner}
      \xdef\total{\total}
      \foreach \val [count=\cv] in \counts {
        \pgfmathsetmacro{\nexttotal}{\total pt+\am/100*\val/5}
        \filldraw[hist \cv]
        (\angle+\delta/2:\total pt)
        arc(\angle+\delta/2:\angle-\delta/2:\total pt)
        -- (\angle-\delta/2:\nexttotal pt)
        arc(\angle-\delta/2:\angle+\delta/2:\nexttotal pt)
        -- cycle;
        \xdef\total{\nexttotal}
        \typeout{\val:\total}
      }
      \pgfmathtruncatemacro{\anglenode}{\angle}
      \ifthenelse{\( \anglenode > 90 \) \AND \( \anglenode < 270 \)}{ 
        \node[rotate=180+\anglenode,anchor=east] at (\angle:\outer) {\id};
      }{
        \node[rotate=\anglenode,anchor=west] at (\angle:\outer) {\id};
      }
    }
  }
\end{tikzpicture}

&

\begin{tikzpicture}[text=black,font=\ttfamily\bfseries]
  \fill[white] circle(\outer+1cm);
  \foreach \id/\counts [count=\cp] in {%
    brex03/{6,(986-6),(1000-986)},
    brex04/{5,(953-5),(1000-953)},
    brex05/{59,(987-59),(1000-987)},
    brex06/{3,(687-3),(1000-687)},
    brex07/{9,(636-9),(1000-636)},
    clic01/{81,(932-81),(1000-932)},
    clic02/{5,(986-5),(1000-986)},
    clic03/{6,(970-6),(1000-970)},
    clic04/{4,(829-4),(1000-829)},
    clic05/{12,(921-12),(1000-921)},
    clic06/{4,(832-4),(1000-832)},
    clic07/{13,(684-13),(1000-684)},
    usel01/{7,(963-7),(1000-963)},
    usel02/{5,(969-5),(1000-969)},
    usel03/{3,(843-3),(1000-843)},
    usel04/{2,(728-2),(1000-728)},
    usel05/{0,(726-0),(1000-726)},
    usel06/{52,(856-52),(1000-856)},
    usel07/{50,(725-50),(1000-725)},
    usel08/{12,(641-12),(1000-641)},
    usel09/{20,(798-20),(1000-798)},
    usel10/{10,(817-10),(1000-817)},
    brex01/{37,(682-37),(1000-682)},
    brex02/{6,(987-6),(1000-987)},
  }{
    \ifthenelse{\equal{\id}{}}{}{
      \pgfmathsetmacro{\angle}{(\cp-1)*\delta-90+\delta/2}
      \pgfmathsetmacro{\total}{\inner}
      \pgfmathsetmacro{\am}{\outer-\inner}
      \xdef\total{\total}
      \foreach \val [count=\cv] in \counts {
        \pgfmathsetmacro{\nexttotal}{\total pt+\am/100*\val/5}
        \filldraw[hist \cv]
        (\angle+\delta/2:\total pt)
        arc(\angle+\delta/2:\angle-\delta/2:\total pt)
        -- (\angle-\delta/2:\nexttotal pt)
        arc(\angle-\delta/2:\angle+\delta/2:\nexttotal pt)
        -- cycle;
        \xdef\total{\nexttotal}
        \typeout{\val:\total}
      }
      \pgfmathtruncatemacro{\anglenode}{\angle}
      \ifthenelse{\( \anglenode > 90 \) \AND \( \anglenode < 270 \)}{ 
        \node[rotate=180+\anglenode,anchor=east] at (\angle:\outer) {\id};
      }{
        \node[rotate=\anglenode,anchor=west] at (\angle:\outer) {\id};
      }
    }
  }
\end{tikzpicture}
\\
(a) & (b)\\[.3cm]
\end{tabular}
\caption{Effectiveness in disarming the obfuscation attack. These radial histograms show the proportion of obfuscated comments correctly scored with at least the same toxicity of the original comment, out of 1000 variants (red area), as obtained by GP model (green sector) and TP+GP method (amber sector). (a) On the \texttt{obfuscation-50} dataset. (b) On the  \texttt{obfuscation-99} dataset. }
\label{fig:polar-hist}
\end{figure}
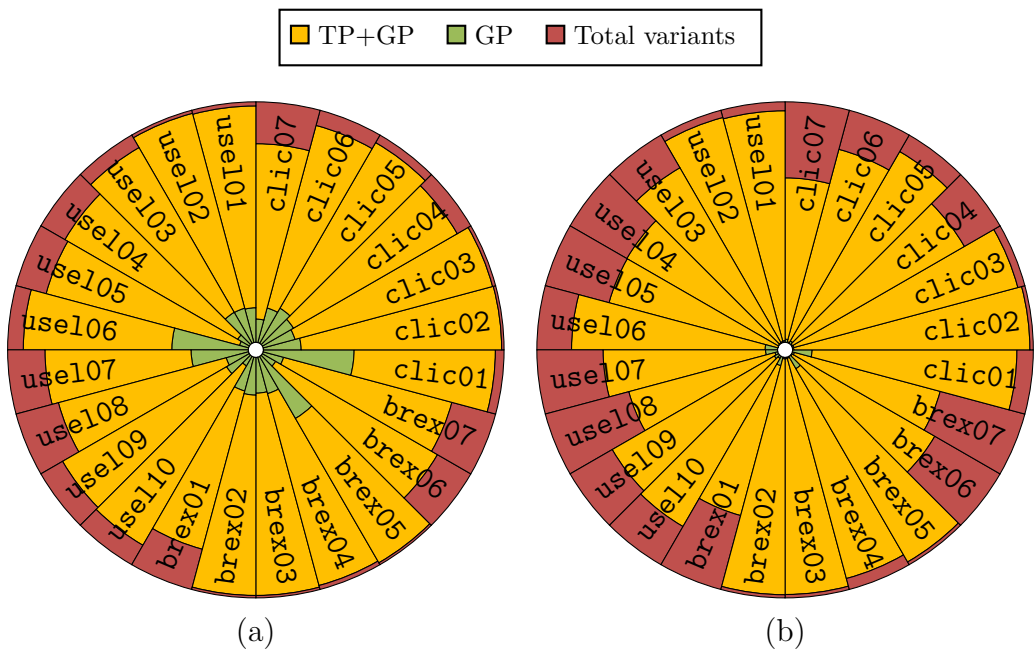

Additionally, it is interesting to note that on the \texttt{obfuscation-50} dataset the GP model (left panel, green area) is able to recognise and score correctly a small proportion of the variants, particularly in the \texttt{usel06}, \texttt{brex05} and \texttt{clic01} categories. On further examination we found that the toxicity of the two former comes from the use of the insulting vocable ``\emph{moron}'' (see \nameref{sup:comments}). Now notice that the homoglyphs defined for this particular combination of letters were simply their upper-- and lowercase versions (see Fig.\ref{fig:homoglyphs}); thus, given a low  corruption rate it is very likely that no segmentators are inserted and therefore some of the generated variants would consist of the same text spelled with intermixed case (e.g. ``\emph{mORoN}'', ``\emph{MoROn}'', etc). In such instances,  the attack can be easily disarmed by a simple case--converting operation which we assume the GP API is carrying out before feeding the comment to the model. 

Regarding the recognition ability of the  \texttt{clic01} variants category by the GP Model, a closer inspection shows that its toxicity is originated from using the term ``\emph{ass}'' in its offensive meaning (see \nameref{sup:comments}); although in this case homoglyphs for these combination of letters include non-letter symbols (see Fig.\ref{fig:homoglyphs}), we observe that because of the short length of the offensive term (3 character-long) along with the low corruption rate (50\%) it is very likely that a number of generated variants are identical to the original or a mixed-case version as before. On the other hand, in the more acid \texttt{obfuscation-99} attacks (right panel) harder obfuscated variants will be found more frequently, hence the dilution of the green bars.   

Moreover, we observe that the longer the toxic sequence and the more toxic terms are included in the original comment, the stronger the deception effect of the obfuscation attack. Take for example \texttt{usel10} which contains 3 toxic terms, one of which is 12-letters long (see \nameref{sup:comments}); in this case, the proportion of recognised variants by the GP model is tiny. Combine that with more frequent bogus segmentations due to higher corruption rates, resulting in the toxic recognition ability of the GP model to further weaken\footnote{Although it is well-worth noting that GP keeps refining its model by continuously learning from more examples, thus fortifying toxic scores appropriately. In fact, the toxicities recorded when this study began, soared up during the following months (see \nameref{sup:toxicities}). Here we report updated scores at the time of submission.}, as it is noticeable in the radial histogram of the \texttt{obfuscation-99} dataset. The TP+GP approach in contrast, is able to still recognise most of them despite the intensity of the attack.

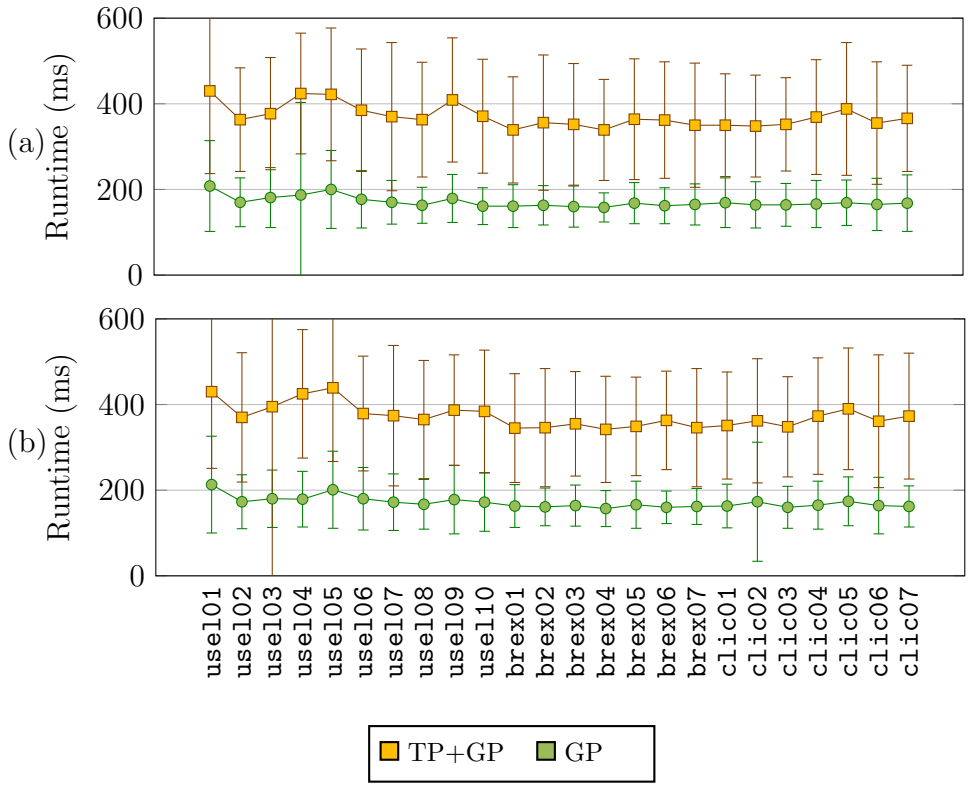
\begin{figure}[t!]
\begin{tikzpicture} 
\begin{axis}[
        title = (a),
        title style = {at={(-.12,0.45)},anchor=east},
        width  = 0.9*\textwidth,
        height = 5cm,
        major x tick style = transparent,
        ymajorgrids = true,
        ymin=0, ymax=600,,
        ylabel = {Runtime (ms)},
        ylabel near ticks,
        xtick=\empty,
        symbolic x coords = {usel01,usel02,usel03,usel04,usel05,usel06,usel07,usel08,usel09,usel10, brex01,brex02,brex03,brex04,brex05,brex06,brex07, clic01,clic02,clic03,clic04,clic05,clic06,clic07},
        xticklabel style = {
            rotate=90,            font={\ttfamily\small}
        },
        enlarge x limits=0.08]
        
        \addplot+[
            mark=*,
            mark options={solid, fill=mygreen},
            color=green!50!black,
            error bars/.cd, 
            y dir=both, y explicit]
        coordinates { 
        (usel01,208) +- (0,106) 
        (usel02,170) +- (0,57) 
        (usel03,181) +- (0,70) 
        (usel04,187) +- (0,216) 
        (usel05,200) +- (0,91) 
        (usel06,177) +- (0,67) 
        (usel07,170) +- (0,51) 
        (usel08,163) +- (0,42) 
        (usel09,179) +- (0,56) 
        (usel10,161) +- (0,43) 
        (brex01,161) +- (0,50) 
        (brex02,163) +- (0,46) 
        (brex03,160) +- (0,48) 
        (brex04,158) +- (0,34) 
        (brex05,168) +- (0,48) 
        (brex06,162) +- (0,42) 
        (brex07,165) +- (0,48) 
        (clic01,169) +- (0,58) 
        (clic02,164) +- (0,54) 
        (clic03,164) +- (0,50) 
        (clic04,166) +- (0,55) 
        (clic05,169) +- (0,53) 
        (clic06,165) +- (0,61) 
        (clic07,168) +- (0,66) 
        };

        \addplot+[
            mark=square*,
            mark options={solid, fill=myamber},
            color=orange!50!black,
            error bars/.cd, 
            y dir=both, y explicit]
        coordinates { 
        (usel01,430) +- (0,193) 
        (usel02,363) +- (0,121) 
        (usel03,377) +- (0,131) 
        (usel04,424) +- (0,141) 
        (usel05,422) +- (0,155) 
        (usel06,385) +- (0,143) 
        (usel07,370) +- (0,173) 
        (usel08,363) +- (0,134) 
        (usel09,409) +- (0,145) 
        (usel10,371) +- (0,133)
        (brex01,339) +- (0,124) 
        (brex02,356) +- (0,158) 
        (brex03,352) +- (0,142) 
        (brex04,339) +- (0,118) 
        (brex05,364) +- (0,141) 
        (brex06,362) +- (0,136) 
        (brex07,350) +- (0,145) 
        (clic01,350) +- (0,120) 
        (clic02,348) +- (0,119) 
        (clic03,352) +- (0,109) 
        (clic04,369) +- (0,134) 
        (clic05,388) +- (0,155) 
        (clic06,355) +- (0,143) 
        (clic07,366) +- (0,124) 
        };
\end{axis} 
\end{tikzpicture}

\begin{tikzpicture}
\begin{axis}[
        title = (b),
        title style = {at={(-.12,0.45)},anchor=east},
        width  = 0.9*\textwidth,
        height = 5cm,
        major x tick style = transparent,
        ymajorgrids = true,
        ymin=0, ymax=600,,
        ylabel = {Runtime (ms)},
        ylabel near ticks,
        xtick=data,
        symbolic x coords = {usel01,usel02,usel03,usel04,usel05,usel06,usel07,usel08,usel09,usel10, brex01,brex02,brex03,brex04,brex05,brex06,brex07, clic01,clic02,clic03,clic04,clic05,clic06,clic07},
        xticklabel style = {
            rotate=90,            font={\ttfamily\small}
        },
        enlarge x limits=0.08]
        
        \addplot+[
            mark=*,
            mark options={solid, fill=mygreen},
            color=green!50!black,
            error bars/.cd, 
            y dir=both, y explicit]
        coordinates { 
        (usel01,213) +- (0,113) 
        (usel02,173) +- (0,63) 
        (usel03,180) +- (0,67) 
        (usel04,179) +- (0,65) 
        (usel05,201) +- (0,90) 
        (usel06,180) +- (0,73) 
        (usel07,172) +- (0,66) 
        (usel08,167) +- (0,58) 
        (usel09,178) +- (0,80) 
        (usel10,172) +- (0,68) 
        (brex01,163) +- (0,50) 
        (brex02,161) +- (0,44) 
        (brex03,164) +- (0,48) 
        (brex04,157) +- (0,42) 
        (brex05,166) +- (0,55) 
        (brex06,160) +- (0,38) 
        (brex07,162) +- (0,42) 
        (clic01,163) +- (0,51) 
        (clic02,173) +- (0,139) 
        (clic03,160) +- (0,49) 
        (clic04,165) +- (0,56) 
        (clic05,174) +- (0,57) 
        (clic06,164) +- (0,66) 
        (clic07,162) +- (0,48) 
        };

        \addplot+[
            mark=square*,
            mark options={solid, fill=myamber},
            color=orange!50!black,
            error bars/.cd, 
            y dir=both, y explicit]
        coordinates { 
        (usel01,430) +- (0,179) 
        (usel02,370) +- (0,151) 
        (usel03,395) +- (0,445) 
        (usel04,425) +- (0,150) 
        (usel05,439) +- (0,172) 
        (usel06,379) +- (0,134) 
        (usel07,374) +- (0,164) 
        (usel08,365) +- (0,138) 
        (usel09,387) +- (0,129) 
        (usel10,384) +- (0,143) 
        (brex01,345) +- (0,127) 
        (brex02,346) +- (0,138) 
        (brex03,355) +- (0,122) 
        (brex04,342) +- (0,124) 
        (brex05,349) +- (0,115) 
        (brex06,363) +- (0,115) 
        (brex07,346) +- (0,138) 
        (clic01,351) +- (0,125) 
        (clic02,362) +- (0,145) 
        (clic03,348) +- (0,117) 
        (clic04,373) +- (0,136) 
        (clic05,390) +- (0,142) 
        (clic06,361) +- (0,155) 
        (clic07,373) +- (0,147) 
        };
\end{axis} 
\end{tikzpicture}

\begin{center}
\begin{tikzpicture}
\node[draw=black,thick,rounded corners=0pt,below=14mm,font=\footnotesize] at (clic.south) {
\begin{tabular}{@{}r@{ }cr@{ }cr@{ }c}
 
 \tikz{\node[draw=black,thick,fill=myamber] (0,0){};} & TP+GP &
 
 \tikz{\node[draw=black,thick,fill=mygreen] (0,0){};}  & GP  &
 
\end{tabular}};
\end{tikzpicture} \\[.2cm]
\end{center}

\caption{Average runtimes for the obfuscation attack experiment, GP model (green) vs. TP+GP method (amber). (a) \texttt{obfuscation-50}. (b)   \texttt{obfuscation-99}. }
\label{fig:runtime}
\end{figure}

Let us focus now on processing time. Fig. \ref{fig:runtime} shows average runtimes of the 1000 repetitions for GP model and TP+GP method in each category. The first observation here is that the effectiveness of the proposed method in disarming the obfuscation attack, comes at the expense of an increase in runtime. This is due of course to the additional preprocessing step that executes the deobfuscation filter. The trend in the plot is that the TP+GP nearly doubles the time taken by GP model alone. 

The second observation is that the average runtimes for each comment category are quite similar in both \texttt{obfuscation-50} and \texttt{obfuscation-99}, a reasonable behaviour considering that the computational complexity of the deobfuscation method depends only on the length of the text \cite{Rojas2017}. Since the obfuscation attack may add extra characters (bogus segmentators or fake letter repetitions) only to the toxic terms within the comment, the total text length may vary but not drastically. Hence, the resemblance  of average runtimes patterns exhibited in both datasets.   

The last observation is related to the variability of the results, indicated by the whiskers in the plots. At first glance these deviations may seem large; nonetheless recall that the experimental pipeline was built using two web-service APIs, one for GP and another one for TP, respectively. In such setting the execution of the experiments need to take into account network latencies and server availability, which are less certain than having a dedicated host running the scripts\footnote{In some cases the GP web service replied with a 400 Bad Request or 502 Bad Gateway error codes; these cases were excluded from the reported average runtimes (651 in \texttt{obfuscation-50} and 506 in \texttt{obfuscation-99} out of their corresponding 24,000 variants).}. Besides, since in our experimental setup the TP+GP method is actually invoking two independent web services, it is more likely to suffer further delays that affect its runtime variability.

\subsection{Polarity attack}
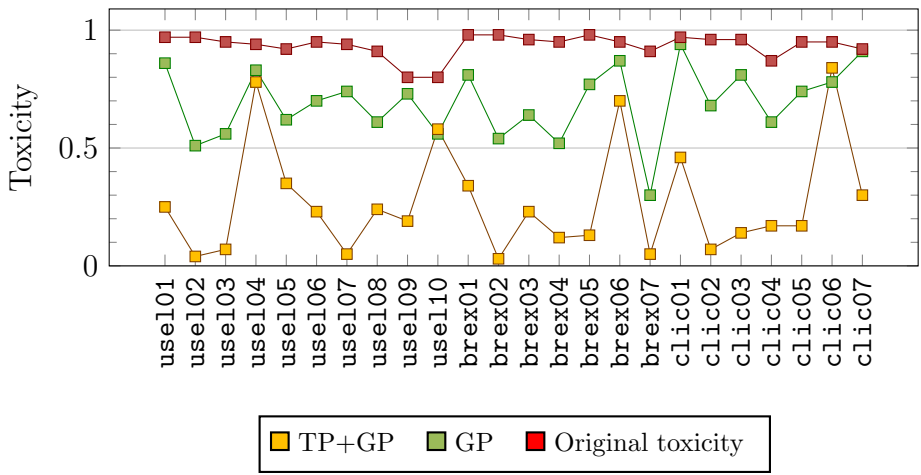
\begin{figure}[t!]
\begin{tikzpicture}
\begin{axis}[
        width  = 0.9*\textwidth,
        height = 5cm,
        minor x tick style = transparent,
        ymajorgrids = true,
        minor tick num=4,
        ymin=0, ymax=1.09,
        ylabel = {Toxicity},
        ylabel near ticks,
        xtick=data,
        symbolic x coords = {usel01,usel02,usel03,usel04,usel05,usel06,usel07,usel08,usel09,usel10, brex01,brex02,brex03,brex04,brex05,brex06,brex07, clic01,clic02,clic03,clic04,clic05,clic06,clic07},
        xticklabel style = {
            rotate=90,            font={\ttfamily\small}
        },
        enlarge x limits=0.08]
        
        \addplot+[
            mark=square*,
            mark options={solid, fill=mygreen},
            color=green!50!black]
        coordinates { 
        (usel01,.86)  
        (usel02,.51) 
        (usel03,.56) 
        (usel04,.83) 
        (usel05,.62) 
        (usel06,.70) 
        (usel07,.74) 
        (usel08,.61) 
        (usel09,.73) 
        (usel10,.56) 
        (brex01,.81) 
        (brex02,.54) 
        (brex03,.64) 
        (brex04,.52) 
        (brex05,.77) 
        (brex06,.87) 
        (brex07,.30) 
        (clic01,.94) 
        (clic02,.68) 
        (clic03,.81) 
        (clic04,.61) 
        (clic05,.74) 
        (clic06,.78) 
        (clic07,.91) 
        };

        \addplot+[
            mark=square*,
            mark options={solid, fill=myamber},
            color=orange!50!black]
        coordinates { 
        (usel01,.25)  
        (usel02,.04) 
        (usel03,.07) 
        (usel04,.78) 
        (usel05,.35) 
        (usel06,.23) 
        (usel07,.05) 
        (usel08,.24) 
        (usel09,.19) 
        (usel10,.58) 
        (brex01,.34) 
        (brex02,.03) 
        (brex03,.23) 
        (brex04,.12) 
        (brex05,.13) 
        (brex06,.70) 
        (brex07,.05) 
        (clic01,.46) 
        (clic02,.07) 
        (clic03,.14) 
        (clic04,.17) 
        (clic05,.17) 
        (clic06,.84) 
        (clic07,.30) 
          };
 
        \addplot+[
            mark=square*,
            mark options={solid, fill=myred},
            color=red!50!black]
        coordinates { 
        (usel01,.97)  
        (usel02,.97) 
        (usel03,.95) 
        (usel04,.94) 
        (usel05,.92) 
        (usel06,.95) 
        (usel07,.94) 
        (usel08,.91) 
        (usel09,.80) 
        (usel10,.80) 
        (brex01,.98) 
        (brex02,.98) 
        (brex03,.96) 
        (brex04,.95) 
        (brex05,.98) 
        (brex06,.95) 
        (brex07,.91) 
        (clic01,.97) 
        (clic02,.96) 
        (clic03,.96) 
        (clic04,.87) 
        (clic05,.95) 
        (clic06,.95) 
        (clic07,.92) 
        };
          
\end{axis} 
\end{tikzpicture}

\begin{center}
\begin{tikzpicture}
\node[draw=black,thick,rounded corners=0pt,below=14mm,font=\footnotesize] at (clic.south) {
\begin{tabular}{@{}r@{ }cr@{ }cr@{ }c}
 
 \tikz{\node[draw=black,thick,fill=myamber] (0,0){};} & TP+GP &
 
 \tikz{\node[draw=black,thick,fill=mygreen] (0,0){};}  & GP  &

 \tikz{\node[draw=black,thick,fill=red] (0,0){};} & Original toxicity  
\end{tabular}};
\end{tikzpicture} \\[.2cm]
\end{center}

\caption{Toxicity scores for the polarity attack, as obtained by GP model (green) vs. TP+GP method (amber) in the 24 negated comments of the \texttt{polarity} dataset. Toxicities of the original comments are also shown (red).}
\label{fig:polarity}
\end{figure}

The results of the polarity attack experiment are summarised in Fig. \ref{fig:polarity}, were toxicities for the original comments and the negated variants using both GP and TP+GP are shown; recall that the \texttt{polarity} dataset contains a single negated variation per comment, hence the figure reports a single run per category  (\{\texttt{usel01',\ldots,clic07'}\}).  These scores indicate that this attack also manages to deceive the GP model, because the negated versions were assigned  high toxicities (although lower than their respective original comments). When using the TP+GP method on the contrary, the toxicities fall to levels mostly below 0.5, except for some comments that incidentally contain toxic content not initially  identified as such at the beginning of our study, but incorporated into newer versions of the GP model (e.g. terms like ``\emph{dishonest}'' in \texttt{usel04} or ``\emph{short-sighted}'' in \texttt{clic06}, that do not appear in the vocabulary lists of Fig. \ref{fig:method}). 

The rationale of this behaviour is that the toxicity model is trained on the assumption that the comments would be coherently written, in a grammatical sense. Thus it simply checks weather toxicity features and patterns are identified within the comment and not if it is correctly constructed. To illustrate this point, let us analyse the effect of the neutralise filter on comment \texttt{usel01'}: \emph{If they voted for Hilary they are NOT idiots}. Using GP, this comment scores a 86\% toxicity score (very close to the score of the original aggressive comment), which when removing the negated offensive term (\emph{If they voted for Hilary they are }) plunges to a 25\% toxicity level; in other words, GP would identify the trimmed sentence as healthy, despite its dubious grammar soundness. 

The TP+GP approach takes advantage of the aforementioned assumption, by simply removing subsequences matching the list of negated predicates. One may argue however, that the trimmed comment is not comparable with the negated one, as the former may become a grammatically malformed sentence (due to the removed predicates). The reason we proceeded that way is we wanted to preserve as far as possible the authentic source content, allowing the filter to remove but not to alter or add new content \emph{ad libitum}. An alternative approach would have been precisely to replace, instead of to remove, the matching negated predicates with  non--negated synonyms (e.g. \emph{not awful} $\rightarrow$ \emph{wonderful}, \emph{not idiot} $\rightarrow$ \emph{clever}); as a result, correctly constructed toxic--trigger--free comments would have been obtained. Hence, their scores would also have dropped down to the healthiness region (in this example \emph{If they voted for Hilary they are clever}, obtained a 24\% toxicity, almost equalling the trimmed sentence).

As a closing remark, runtimes of this experiment resembled those of the obfuscation attack in the sense that TP+GP nearly doubled GP processing time (data not shown).

\section{Conclusion}
Machine learning models of language toxicity such as the GP engine, are effectively able to estimate  aggressiveness in written comments. Nonetheless, these models can be deceived by adversarial manipulation of the text, misusing the linguistic assumptions on which they are trained. Firstly at the character--level, an adversary may take advantage of the versatility of written text along with the robust reconstruction capacity of the human brain, to replace similar graphemes and/or to insert fake segmentations yielding toxic contents unrecognisable to the model, thus increasing its false negative rate. Secondly, at the sentence level, an adversary may deceive the model using grammatical operations as simple as negations, switching the polarity of the comment but keeping a structure similar to previously learned toxic examples, hence, increasing the model false positive rate.  

These attacks can be alleviated by preprocessing the comment so as to restore its standard sequence and/or polarity representation before feeding the toxic model. In the first kind of attack, such an approach can be effective even if half the length of the toxic sequence is obfuscated; higher corruption rates decrease to some extent its effectiveness, but then also the deception attempt becomes futile since the obfuscation becomes illegible to human readers. In the second kind of attack, by removing the negated toxic predicates or replacing them with affirmative synonyms, the preprocessor is able to restore the correct toxicity; however we anticipate further research is required since negation is a complex grammatical category whose rules differ amply between lexical elements such as verbs compared to adjectives, nouns and other clauses.

Another avenue of future study may consider harder adversarial attacks such as transposition or tandem--character obfuscations, that despite lo-sing some of the visual deception factor still pose alluring challenges (e.g., \mbox{\texttt{fuck} $\rightarrow$ \texttt{fukc}},  \texttt{SHIT} $\rightarrow$ \texttt{S|-|IT}). Besides, computational speedups in the execution of the filters are also of practical interest (see \cite{Navarro2001} or more recently, \cite{Nicolae2015}), considering that language is an entity in constant evolution\cite{Hodson2016} whose toxic vocabulary evolves likewise. Moreover, it would be feasible to extend our adversarial toxic comments detection to other  Latin--based languages by training their respective word- or character-level $n$-grams machine learning models \cite{Perspective2017}  
coupled with filters of obfuscation--prone customised vocabulary \cite{Rojas2017}. Lastly, exploring feature spaces derived from conversational networks \cite{Papegnies2017} may improve the effectiveness of our methods.

As a final remark, we believe that ensuring aggression--free, respectful and opinionated online discussions would require a pipeline of text processors working on different levels of abstraction, ranging from sequence to linguistic to semantic to sentiment analysis; lower level processing would provide useful features to higher level analysis, which of course, may prove more difficult to handle. To illustrate this point, take for example figures of speech such as sarcasm, irony or satire that may use toxic content as a legitimate rhetorical device. In any case, aggression is strongly related to arising causes of concern such as cyberbullying, harassment, post-truth opinion forming, pedophilia, terrorism and mind state manipulation towards violence \cite{Stone2015, Burnap2016, Hosseinmardi2015}; hence, robust toxicity detection will have significant implications for casual, legal or political issues in online communities.

\bigskip
\noindent\textbf{Disclaimer:} Original comments and modified variants, as depicted in Tables 1 and 3 and in other sections of this paper, were collected or derived from their original source \cite{GPwebsite} for research purposes only; they do not represent the views or opinions of the authors.


\section*{Appendixes}


\subsection*{Appendix A. Original comments}
\label{sup:comments}
Table \ref{tab:comments} shows the text sequences of the original toxic comments extracted from the GP Website \cite{GPwebsite} with their toxicity scores at the beginning of this study. 

\renewcommand{\arraystretch}{1.1}
\begin{table}[t!]
\footnotesize
 \caption{\bf The list of original toxic comments extracted from \cite{GPwebsite}. }
 \begin{tabular}{|c|m{9cm}|c|c|} 
 \hline
{\bf ID} & {\bf Comment sequence} & {\bf Toxicity} & {\bf Length} \\ [0.5ex] 
 \hline\hline
\texttt{usel01} & If they voted for Hilary they are \textbf{idiots} & 0.90 & 40 \\\hline
\texttt{usel02} & It was \textbf{awful}. People are \textbf{stupid}. & 0.90 & 32 \\\hline
\texttt{usel03} & I respect it but they are \textbf{stupid} & 0.89 & 32 \\\hline
\texttt{usel04} & Liberals are \textbf{idiots}. Why in the \textbf{hell} would u vote for a person like Hillary Clinton. She believes in stealing, she's dishonest and believes it's okay for an \textbf{abortion} to be done at 9 months. She's \textbf{awful}. & 0.87 & 202 \\\hline
\texttt{usel05} & Donald trump is the \textbf{worst} person to be president. This election really showed how \textbf{stupid} the U.S. is. & 0.85 & 101 \\\hline
\texttt{usel06} & Anyone who voted for Trump is a \textbf{moron}. & 0.81 & 38 \\\hline
\texttt{usel07} & \textbf{Screw} you trump supporters & 0.79 & 26 \\\hline
\texttt{usel08} & It was a \textbf{terrible}. Both sides \textbf{suck}, but Trump REALLY is scary & 0.72 & 61 \\\hline
\texttt{usel09} & It is a \textbf{shame} that Donald Trump was elected. You can never underestimate the \textbf{stupidity} of middle America. & 0.67 & 105 \\\hline
\texttt{usel10} & Good job voting for a \textbf{racist}, \textbf{sexist} guy who's putting white \textbf{supremacists} in the white house. Have fun. & 0.67 & 103 \\\hline\hline
\texttt{brex01} & Anti democratic \textbf{assholes} & 0.95 & 24 \\\hline
\texttt{brex02} & \textbf{Ignorant} and \textbf{stupid} & 0.93 & 19 \\\hline
\texttt{brex03} & They are \textbf{stupid} and \textbf{ignorant} with no class & 0.91 & 42 \\\hline
\texttt{brex04} & It's \textbf{stupid} and wrong & 0.89 & 21 \\\hline
\texttt{brex05} & you \textbf{morons} & 0.92 & 10 \\\hline
\texttt{brex06} & \textbf{idiots}. backward thinking people. \textbf{nationalists}. not accepting facts. susceptible to lies. & 0.80 & 89 \\\hline
\texttt{brex07} & It's \textbf{rubbish} & 0.69 & 12 \\\hline\hline
\texttt{clic01} & They have their heads up their \textbf{ass}. & 0.93 & 35 \\\hline
\texttt{clic02} & How can you be so \textbf{stupid}? & 0.91 & 25 \\\hline
\texttt{clic03} & They are liberal \textbf{idiots} who are \textbf{uneducated}. & 0.90 & 43 \\\hline
\texttt{clic04} & They're \textbf{stupid}, it's getting warmer, we should enjoy it while it lasts. & 0.86 & 71 \\\hline
\texttt{clic05} & Climate change is happening and it's not changing in our favor. If you think differently you're an \textbf{idiot}. & 0.84 & 105 \\\hline
\texttt{clic06} & I think those people are \textbf{stupid} and short-sighted & 0.84 & 49 \\\hline
\texttt{clic07} & They're allowed to do that. But if they act like \textbf{assholes} about, I will block them. & 0.78 & 83 \\
 \hline
 \end{tabular}
 \label{tab:comments}
\end{table}

\clearpage
\subsection*{Appendix B. Datasets and experiment source code}
\label{sup:datasets}
The \texttt{obfuscation-50},  \texttt{obfuscation-99}, and \texttt{polarity} datasets as well as the \texttt{Go} language source code used in our experiments are publicly available at the following Gitlab repository:  \url{https://gitlab.com/textpatrol/gp-tp-experiment}.

\subsection*{Appendix C. GP Model toxicity update}
\label{sup:toxicities}
\begin{figure}[!ht]
\begin{tikzpicture} 
\begin{axis}[
        title = (a),
        title style = {at={(-.12,0.45)},anchor=east},
        width  = 0.9*\textwidth,
        height = 3.5cm,
        major x tick style = transparent,
        ymajorgrids = true,
        ymin=0., ymax=1.05,
        ylabel = {Toxicity},
        ylabel near ticks,
        xtick=\empty,
        symbolic x coords = {usel01,usel02,usel03,usel04,usel05,usel06,usel07,usel08,usel09,usel10, brex01,brex02,brex03,brex04,brex05,brex06,brex07, clic01,clic02,clic03,clic04,clic05,clic06,clic07},
        xticklabel style = {
            rotate=90,            font={\ttfamily\small}
        },
        enlarge x limits=0.08]
        
        \addplot+[
            mark=square*,
            mark options={solid, fill=mygreen},
            color=green!50!black]
        coordinates { 
        (usel01,.90)  
        (usel02,.90) 
        (usel03,.89) 
        (usel04,.87) 
        (usel05,.85) 
        (usel06,.81) 
        (usel07,.79) 
        (usel08,.72) 
        (usel09,.67) 
        (usel10,.67) 
        (brex01,.95) 
        (brex02,.93) 
        (brex03,.91) 
        (brex04,.89) 
        (brex05,.92) 
        (brex06,.80) 
        (brex07,.69) 
        (clic01,.93) 
        (clic02,.91) 
        (clic03,.90) 
        (clic04,.86) 
        (clic05,.84) 
        (clic06,.84) 
        (clic07,.78) 
        };

        \addplot+[
            mark=square*,
            mark options={solid, fill=myamber},
            color=orange!50!black]
        coordinates { 
        (usel01,.99)  
        (usel02,.94) 
        (usel03,.93) 
        (usel04,.97) 
        (usel05,.96) 
        (usel06,.95) 
        (usel07,.90) 
        (usel08,.87) 
        (usel09,.83) 
        (usel10,.80) 
        (brex01,.97) 
        (brex02,.94) 
        (brex03,.97) 
        (brex04,.88) 
        (brex05,.96) 
        (brex06,.95) 
        (brex07,.64) 
        (clic01,.96) 
        (clic02,.97) 
        (clic03,.97) 
        (clic04,.92) 
        (clic05,.94) 
        (clic06,.98) 
        (clic07,.95) 
          };
 
\end{axis} 
\end{tikzpicture}

\begin{tikzpicture}
\begin{axis}[
        title = (b),
        title style = {at={(-.12,0.45)},anchor=east},
        width  = 0.9*\textwidth,
        height = 3.5cm,
        major x tick style = transparent,
        ymajorgrids = true,
        ymin=0., ymax=1.05,
        ylabel = {Toxicity},
        ylabel near ticks,
        xtick=data,
        symbolic x coords = {usel01,usel02,usel03,usel04,usel05,usel06,usel07,usel08,usel09,usel10, brex01,brex02,brex03,brex04,brex05,brex06,brex07, clic01,clic02,clic03,clic04,clic05,clic06,clic07},
        xticklabel style = {
            rotate=90,            font={\ttfamily\small}
        },
        enlarge x limits=0.08]
        
        \addplot+[
            mark=square*,
            mark options={solid, fill=blue},
            color=blue!50!black]
        coordinates { 
        (usel01,.90)  
        (usel02,.87) 
        (usel03,.84) 
        (usel04,.79) 
        (usel05,.72) 
        (usel06,.80) 
        (usel07,.71) 
        (usel08,.71) 
        (usel09,.57) 
        (usel10,.57) 
        (brex01,.94) 
        (brex02,.92) 
        (brex03,.83) 
        (brex04,.83) 
        (brex05,.88) 
        (brex06,.84) 
        (brex07,.72) 
        (clic01,.91) 
        (clic02,.87) 
        (clic03,.85) 
        (clic04,.59) 
        (clic05,.83) 
        (clic06,.81) 
        (clic07,.78) 
        };

        \addplot+[
            mark=square*,
            mark options={solid, fill=red},
            color=red!50!black]
        coordinates { 
        (usel01,.86)  
        (usel02,.51) 
        (usel03,.56) 
        (usel04,.83) 
        (usel05,.62) 
        (usel06,.70) 
        (usel07,.74) 
        (usel08,.61) 
        (usel09,.73) 
        (usel10,.56) 
        (brex01,.81) 
        (brex02,.54) 
        (brex03,.64) 
        (brex04,.52) 
        (brex05,.77) 
        (brex06,.87) 
        (brex07,.30) 
        (clic01,.94) 
        (clic02,.68) 
        (clic03,.81) 
        (clic04,.61) 
        (clic05,.74) 
        (clic06,.78) 
        (clic07,.91) 
        };
          
\end{axis} 
\end{tikzpicture}

\begin{center}
\begin{tikzpicture}
\node[draw=black,thick,rounded corners=0pt,below=14mm,font=\footnotesize] at (clic.south) {
\begin{tabular}{@{}r@{ }r@{ }cr@{ }cr@{ }r@{ }cr@{ }c}
 
 (a) Original: & \tikz{\node[draw=black,thick,fill=myamber] (0,0){};} & Sep. &
 
 \tikz{\node[draw=black,thick,fill=mygreen] (0,0){};}  & Feb. &

 (b) Negated: &
 \tikz{\node[draw=black,thick,fill=red] (0,0){};} & Sep. &
 
  \tikz{\node[draw=black,thick,fill=blue] (0,0){};} & Feb.
  
\end{tabular}};
\end{tikzpicture} \\[.2cm]
\end{center}

\caption{GP toxicity scores variation during the lasts months (Feb-Sep/2017).}
\label{fig:toxicities}
\end{figure}
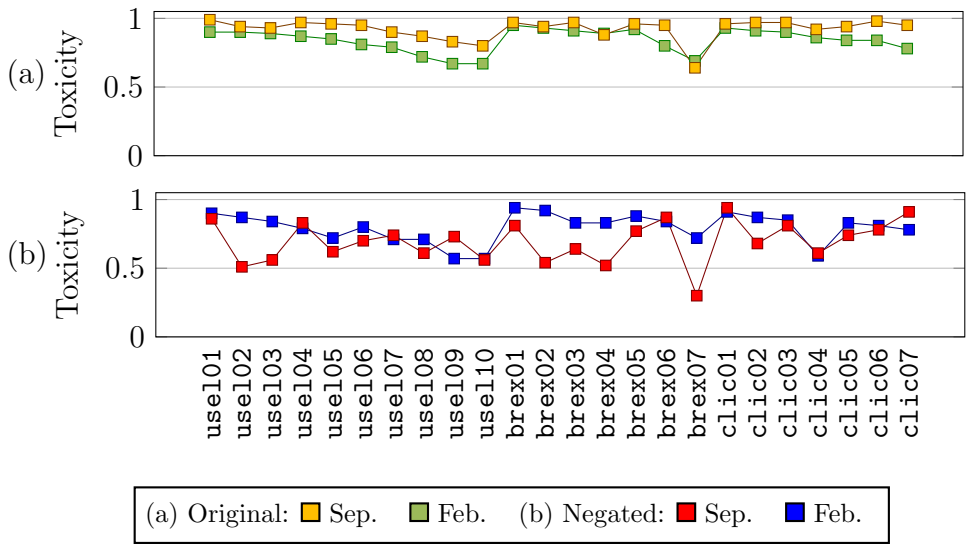

Fig. \ref{fig:toxicities} shows a comparative plot of the toxicity scores given by the GP model at the beginning of the study (February, 2017) versus at the time of submission (September, 2017). For the original comments it can be seen that scores have raised, suggesting that the toxicity model has been refined during this period, probably by learning from more examples and newer toxic contents. On the contrary, the negated comments show a trend of decreasing toxicities; we believe the model has been refined to account for additional linguistic features related to the grammatical structures of negation, although still it is being deceived by this attack to obtain scores higher than 50\%. 

\end{document}